%% file: eamt24.tex
\pdfoutput=1

\documentclass[11pt]{article}
\usepackage{eamt24}
\usepackage{times}
\usepackage{url}
\usepackage{latexsym}
\usepackage[small,bf]{caption}
\usepackage{microtype}
\usepackage{inconsolata}
\usepackage{booktabs}
\usepackage{placeins}
\usepackage{enumitem}
\usepackage{multicol}
\usepackage{multirow}
\usepackage{makecell}
\usepackage{arydshln}
\usepackage{tabularx}
\usepackage{CJKutf8}
\usepackage{subcaption}

\usepackage{dblfloatfix}
\usepackage{float}

\usepackage{tikz}
\usepackage{pgfplots}
\usepackage{pdflscape}
\usepackage{pgfplotstable}
\pgfplotsset{
    compat=1.17,
    every non boxed x axis/.append style={x axis line style=-},
    every non boxed y axis/.append style={y axis line style=-},
}
\usetikzlibrary{patterns}
\usetikzlibrary{matrix}
\usepgfplotslibrary{groupplots}

\newcommand{\source}[1]{\textcolor{orange}{#1}}
\newcommand{\translation}[1]{\textcolor{olive}{#1}}
\newcommand{\randomtranslation}[1]{\textcolor{cyan}{#1}}
\newcommand{\lang}[1]{\textcolor{magenta}{#1}}

\newcommand{\myargmax}{\arg\!\max}

\title{Iterative Translation Refinement with Large Language Models}

\author{
Pinzhen Chen\textsuperscript{1}\quad
Zhicheng Guo\textsuperscript{2}\quad
Barry Haddow\textsuperscript{1}\quad
Kenneth Heafield\textsuperscript{1}\\
\\
\textsuperscript{1}School of Informatics, University of Edinburgh\\
\textsuperscript{2}Dept. of Comp. Sci. \& Tech., Institute for AI, Tsinghua University\\
\texttt{\{pinzhen.chen,bhaddow,kenneth.heafield\}@ed.ac.uk}\\
\texttt{guo-zc21@mails.tsinghua.edu.cn}\\
}

\date{}

\begin{document}
\maketitle
\begin{abstract}
We propose iteratively prompting a large language model to self-correct a translation, with inspiration from their strong language understanding and translation capability as well as a human-like translation approach. Interestingly, multi-turn querying reduces the output's string-based metric scores, but neural metrics suggest comparable or improved quality. Human evaluations indicate better fluency and naturalness compared to initial translations and even human references, all while maintaining quality. Ablation studies underscore the importance of anchoring the refinement to the source and a reasonable seed translation for quality considerations. We also discuss the challenges in evaluation and relation to human performance and translationese.
\end{abstract}

\section{Introduction}
Large language models (LLMs), e.g.~generative pre-trained Transformers (GPT), have made notable advancements in natural language processing \cite{Radford2019Language,brown2020language,kaplan2020scaling,ouyang2022training}. In machine translation (MT), where the convention is to use an encoder-decoder architecture to deal with source and target sentences respectively \cite{Bahdanau2015,vaswani2017}, recent papers have examined the feasibility of LLM prompting for translation \cite{vilar-etal-2023-prompting,zhang2023prompting,hendy2023good,agrawal-etal-2023-context}.

With autoregressive decoding being the convention, machine translation models yield output in a single attempt, and so do post-editing models. Rather, a human translator can read and edit translations repeatedly, or even pass the outcome to another translator for a second opinion. We explore such an iterative refinement process with LLMs, where the proposed method simply feeds a source-translation pair into an LLM for an improved translation in multiple rounds. It is worth noting that this method can be applied to an initial translation from any model, not just LLM outputs. We further conduct a qualitative evaluation of the outputs. Our approach offers two insights from a fluency and naturalness perspective: 1) LLMs are pre-trained on natural texts that are orders of magnitude larger than traditional MT data, and 2) the method does not require complicated prompt engineering, yet allows for iterative and arbitrary rephrasing compared to automatic post-editing, which is limited to token-level error correction without style editing \cite{ive-etal-2020-post}. 

Empirical results show that the refinement procedure introduces significant textual changes reflected by the drop in BLEU and chrF++, but attains similar or higher COMET scores compared to initial translations. Native speakers prefer refined outputs in terms of fluency and naturalness when compared with GPT translations and even human references. Reference-based human evaluation confirms that such gains are made without sacrificing general quality. As corroborated by recent works, automatic metrics like BLEU and COMET are witnessed to move in opposite directions \cite{freitag-etal-2019-ape,freitag-etal-2022-results}. Our human-like LLM prompting method contributes to translation naturalness which can enhance utility as perceived by the target language users. On a broader scope, this work touches on the concept of involving LLMs in a collaborative translation editing strategy.

\begin{table*}[th]
\centering
\small
\input{tables/table_prompt_text.tex}
\caption{Prompts used in our work, where a \texttt{\$\{variable\}} is substituted with its corresponding content.}
\label{tab:prompt_text}
\end{table*}

\section{Methodology}
Having an input source sentence $x$ and an optimizable model $\theta_{mt}$, the process to obtain a translation $y$ can be modelled as \(y=\myargmax_{y} P(y|x;\theta_{mt})\). Next, an automatic post-editor $\theta_{ape}$ creates a refined translation $y'$ through modelling \(y'=\myargmax_{y'} P(y'|x,y;\theta_{ape})\). Conventional translation or automatic post-editing models are trained on \((x,y)\) or \((x,y,y')\) data pairs.

Extending prior work on LLM prompting, our study uses zero-shot prompting by affixing a task description to form a prompt $p$ and querying an LLM $\theta_{LLM}$ to elicit a response \cite{brown2020language}. We introduce five prompts in our study:
\begin{enumerate}[nolistsep] 
    \item \textit{Translate}: it queries for a translation of a source input, extending the translation process with a prompt $p$: \(y=\myargmax_{y} P(y|p,x;\theta_{LLM})\). This is vanilla LLM prompting for MT.
    \item \textit{Refine}: similar to post-editing, the LLM is given the source sentence and the previous translation to produce a better translation \(y'=\myargmax_{y'} P(y'|p,x,y;\theta_{LLM})\).
    \item \textit{Refine}\textsubscript{Contrast}: as a contrasting prompt to the above, we insert the word ``bad'' to hint that the previously translated text is unwanted, regardless of its actual quality.
    \item \textit{Refine}\textsubscript{Random}: same prompt as \textit{Refine}\textsubscript{Contrast}, but in the first iteration, a random sentence is fed instead of a translation to imitate a genuinely ``bad translation''.
    \item \textit{Paraphrase}: a contrasting experiment to translation prompting, we ask an LLM to rephrase a translation without feeding the source sentence $x$: \(y''=\myargmax_{y''} P(y''|p,y;\theta_{LLM})\).
\end{enumerate}
We propose to iteratively call the refinement prompts, where the source stays the same but the previous translation is updated each turn. To encourage a parsable model response, we ask the LLM to not give any explanation. Such prompting does not require model parameters $\theta_{LLM}$ to be accessible. Through ablation prompts, \textit{Refine}\textsubscript{Random} and \textit{Paraphrase}, we analyse to what degree the source input and seed translations are helpful. The exact prompt texts are displayed in Table~\ref{tab:prompt_text}.

\section{Experiments}
\subsection{Data and model details}

We select language pairs from the news and general domain translation tasks hosted at WMT 2021 and 2022 \cite{akhbardeh-etal-2021-findings,kocmi-etal-2022-findings}, which are supported by COMET to obtain reliable scores. In total, we tested seven translation directions: English$\leftrightarrow$German (en$\rightarrow$de, de$\rightarrow$en), English$\leftrightarrow$Chinese (en$\rightarrow$zh, zh$\rightarrow$en), German$\rightarrow$French (de$\rightarrow$fr), English$\rightarrow$Japanese (en$\rightarrow$ja), and Ukrainian$\rightarrow$Czech (uk$\rightarrow$cs). We directly benchmark on the test sets, and in situations where multiple references are available, we use human reference ``A'' released by the WMT organizers as our reference.

We experiment with GPT-3.5, a powerful closed-source model from OpenAI that can be accessed by all users.\footnote{We accessed a version of \texttt{gpt-3.5-turbo} with training data up to Sep 2021, so it should not have seen WMT 2021 or 2022 test references. Nevertheless, our findings are mostly drawn from reference-free metrics and human evaluation.} As the API call tends to be slow, we randomly sample 200 instances from the official test set to form our in-house test. In the refinement and paraphrase experiments, we use the response from the LLM \textit{Translate} query as the seed translation to be improved upon. We do not keep the query (multi-turn) history so as to prevent an LLM from seeing that the previous translation is produced by itself. In experiments later on, we also tested with translations from encoder-decoder systems that participated in WMT, human references, and online systems. Overall, translation refinement is iterated four times at maximum considering the API costs.

\subsection{Evaluation setup}
\label{sec:evaluation-setup}
We consider four automatic metrics: string-based BLEU \cite{papineni-etal-2002-bleu} and chrF++ \cite{popovic-2017-chrf} as well as embedding-based COMET\textsubscript{DA} and COMET\textsubscript{QE} \cite{rei-etal-2020-comet}. The difference between the DA and QE versions is that COMET\textsubscript{DA} requires a source, a translation, and a reference, whereas COMET\textsubscript{QE} is reference-free. BLEU and chrF++ are as implemented in the \texttt{sacrebleu} toolkit.\footnote{\url{https://github.com/mjpost/sacrebleu}} We also use this toolkit to obtain test sets with references as well as past WMT systems' outputs. Specifically for tokenization in BLEU calculation, we use ``zh'' for Chinese, ``ja-mecab'' for Japanese, and ``13a'' for the rest. The BLEU and chrF++ signatures are footnoted.\footnote{\texttt{\#:1|c:mixed|e:no|tok:13a|s:exp|v:2.3.1}}\textsuperscript{,}\footnote{\texttt{\#:1|c:mixed|e:yes|nc:6|nw:2|s:no|v:2.3.1}} For COMET metrics, we used the official implementation released by the authors.\footnote{\url{https://github.com/Unbabel/COMET}} 

\begin{table}[t]
    \centering\small
    \setlength{\tabcolsep}{0.45ex}
    \input{tables/table_automatic_results_gpt_previous.tex}
    \caption{Automatic scores of different strategies with GPT on high-resource pairs from WMT 2021 news translation.}
    \label{tab:automatic_results_gpt_main}
\end{table}

\begin{table}[th]
    \centering\small
    \setlength{\tabcolsep}{0.45ex}
    \input{tables/table_automatic_results_non_en_previous.tex}
    \caption{Automatic scores of different strategies with GPT on low-resource and medium-resource pairs from WMT 2022 news translation.}
    \label{tab:automatic_results_non_en}
\end{table}

\subsection{Refinement results}
\label{sec:gpt_exp}
\paragraph{WMT21} We first experiment with en$\leftrightarrow$de and en$\leftrightarrow$zh from WMT21, which are high-resource languages in terms of both translation data and LLM training data. We run all five prompts 
and display results in Table~\ref{tab:automatic_results_gpt_main}. For iterative refinement and paraphrasing experiments, the best iteration is picked according to COMET\textsubscript{QE}. We observe that the refined translations record a drastic drop in string-based metrics compared to initial translations, indicating lexical and structural variations. In terms of COMET\textsubscript{DA}, refined outputs surpass initial GPT translations in three out of four cases, and in terms of COMET\textsubscript{QE}, the refinement strategy ends as the highest with substantial improvement for into-English directions. As a contrasting experiment, \textit{Paraphrase} sees a decline in all metrics, suggesting the importance of feeding the source input as an anchor during iterations to prevent semantic drift.

\paragraph{WMT22} Moving to lower-resourced languages with non-English translation, we gather numbers for three translation directions from WMT22 in Table~\ref{tab:automatic_results_non_en}. Since \textit{Refine}\textsubscript{Random} results are not desirable for WMT21, we omit experiments with this. The overall pattern remains the same as before: \textit{Refine} works best, obtaining higher COMET\textsubscript{QE} than vanilla translations and \textit{Refine}\textsubscript{Contrast}. Also, the reduction in string-based scores becomes less obvious, which might be attributed to seed GPT translations in lesser-resourced languages being lower in quality in the beginning.

\begin{table}[th]
    \centering\small
    \setlength{\tabcolsep}{0.45ex}
    \input{tables/table_automatic_results_wmt_previous.tex}
    \caption{Automatic scores of refining WMT 2021 news shared task German-to-English submissions.}
    \label{tab:automatic_results_wmt}
\end{table}

\paragraph{Online systems, encoder-decoder systems, and human translations}\label{sec:wmt-refinement} In addition to translation refinement from GPT-3.5 itself, we also apply our refinement calls to outputs from conventional MT systems and human translators. These translations can represent genuine errors, if any, introduced during the translation process. Out of the seven WMT21 submissions, we select outputs from four models built by research labs that, based on human evaluation, have been ranked at significantly different positions on the German-to-English leaderboard: Tencent \cite{wang-etal-2021-tencent}, Facebook AI \cite{tran-etal-2021-facebook}, Edinburgh \cite{chen-etal-2021-university}, and Huawei TSC \cite{wei-etal-2021-hw}. These are competitive systems built with data augmentation, multilingualism, ensembling, re-ranking, etc. We then include two online engines used in WMT 2021: Online-A and Online-Y. Finally, human reference ``B'' is added so that we can experiment with our refinement strategy on human translations.\footnote{The overview paper of WMT 2021 states that ``for German$\leftrightarrow$English, the `B' reference was found to be a post-edited version of one of the participating online systems''. We discover that it refers to English$\rightarrow$German only, and German$\rightarrow$English is not affected.} References ``A'' and ``B'' are sourced from different translation agencies \cite{akhbardeh-etal-2021-findings}.

We report automatic scores from the refinement process in Table~\ref{tab:automatic_results_wmt}. A pattern similar to previous GPT translation refinement is noticed: for five out of seven WMT entries, the refinement strategy reaches a higher COMET\textsubscript{QE} score, surprisingly, with up to one-third drop in BLEU. \textit{Refine}\textsubscript{Contrast} in all but one system surpass \textit{Refine}, and without the initial translation, \textit{Paraphrase} iterations record the lowest scores compared to the original submissions and refinements.

\section{Human Evaluation}
String-based and neural scores are observed to vary in opposite directions, which may suggest volatile changes in texts. Since it is questionable to conclude a quality degradation in this case, we set up human evaluations to measure two characteristics in the refined translations: text naturalness and overall quality. Human evaluators involved in this study are practitioners in the field of natural language processing but are unaware of the goal of this study.

\begin{figure*}[th]
\centering\small
{\begin{minipage}[t]{0.31\textwidth}\centering
\input{plots/plot_source_free_refine_vs_translate_new.tex}
\end{minipage}%
\begin{minipage}[t]{0.31\textwidth}\centering
\input{plots/plot_source_free_refine_vs_ref_new.tex}
\end{minipage}%
\begin{minipage}[t]{0.19\textwidth}\centering
\input{plots/plot_source_based_refine_vs_translate_new.tex}
\end{minipage}%
\begin{minipage}[t]{0.19\textwidth}\centering
\input{plots/plot_source_based_refine_vs_ref_new.tex}
\end{minipage}}
\vspace{-1.5ex}
\input{plots/plot_source_based_free_legend.tex}
\vspace{-1.5ex}
\caption{Human preferences on fluency and naturalness (source-free, left) and overall quality (source-based, right).}\label{fig:human-evaluation}
\end{figure*}
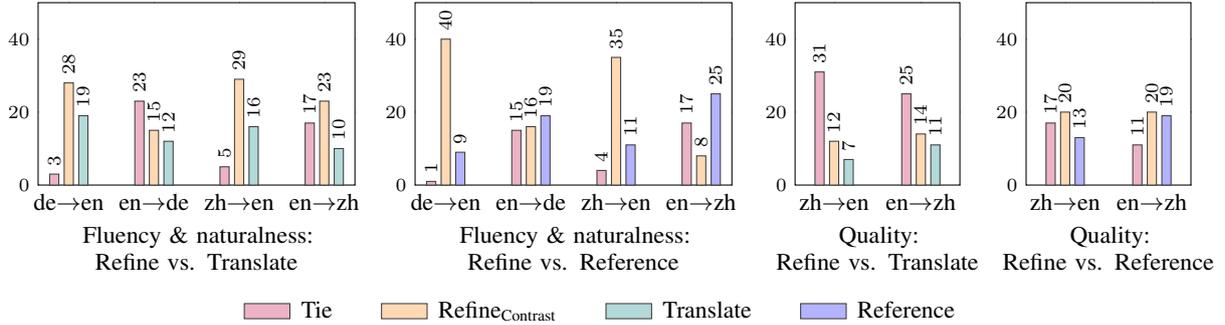

\begin{figure*}[th]
\centering\small
\begin{subfigure}[h]{1\textwidth}
\centering\small
\input{plots/plot_de_en_bleu.tex}\hfill
\input{plots/plot_de_en_comet_da.tex}\hfill
\input{plots/plot_de_en_comet_qe.tex}
\label{fig:de-en}
\end{subfigure}

\begin{subfigure}[h]{1\textwidth}
\centering\small
\input{plots/plot_en_de_bleu.tex}\hfill
\input{plots/plot_en_de_comet_da.tex}\hfill
\input{plots/plot_en_de_comet_qe.tex}
\label{fig:en-de}
\end{subfigure}

\begin{subfigure}[h]{1\textwidth}
\centering\small
\input{plots/plot_zh_en_bleu.tex}\hfill
\input{plots/plot_zh_en_comet_da.tex}\hfill
\input{plots/plot_zh_en_comet_qe.tex}
\label{fig:zh-en}
\end{subfigure}

\begin{subfigure}[h]{\textwidth}
\centering\small
\input{plots/plot_en_zh_bleu.tex}\hfill
\input{plots/plot_en_zh_comet_da.tex}\hfill
\input{plots/plot_en_zh_comet_qe.tex}
\label{fig:en-zh}
\end{subfigure}

\begin{subfigure}[h]{1\textwidth}
\centering\small
\input{plots/plot_legend_scores.tex}
\end{subfigure}

\caption{BLEU, COMET\textsubscript{DA}, and COMET\textsubscript{QE} at different refinement and paraphrase iterations for high-resource translation.}
\label{fig:trends}
\end{figure*}
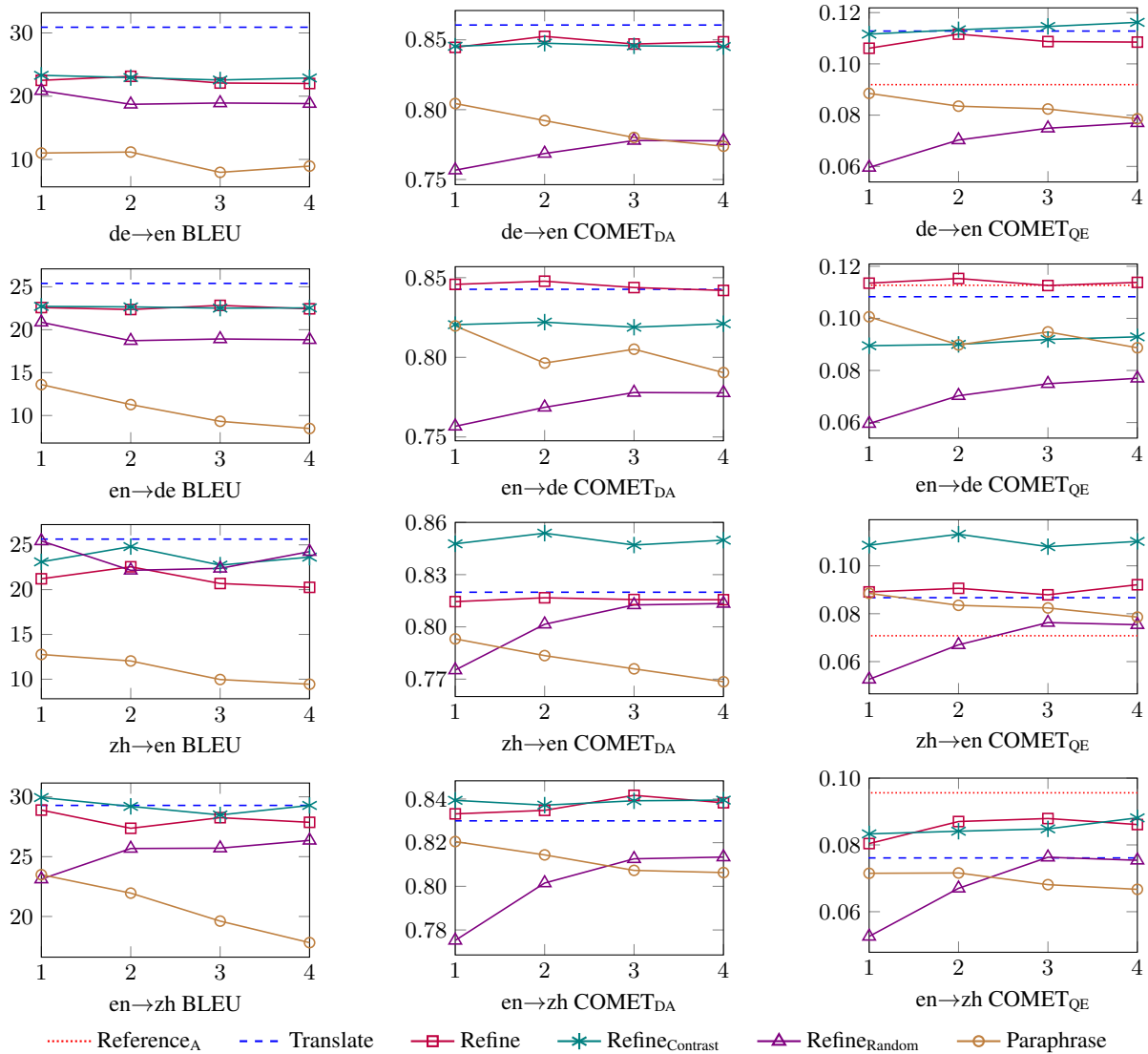

\subsection{Fluency and naturalness}
We mimic the human evaluation of fluency in \cite[p819]{Lembersky-lm}. Native speakers of the target language are with two translations but without the source sentence; then we ask ``\texttt{Please choose the translation that is more fluent, natural, and reflecting better use of {\$\{language\}}}'', where \texttt{\$\{language\}} is substituted with the target language name. The evaluator has three options: they can select one of the two translations, or a ``tie'' if they consider both equally (un)natural. We conduct such pairwise evaluation to compare the first-round output from \textit{Refine}\textsubscript{Contrast} against human references, as well as against \textit{Translate} separately.

We evaluate 50 samples from en$\leftrightarrow$de and en$\leftrightarrow$zh experiments in Section~\ref{sec:gpt_exp}, and report in Figure~\ref{fig:human-evaluation} (left). Native speakers prefer \textit{Refine}\textsubscript{Contrast} to vanilla \textit{Translate} in all four directions, and even favour \textit{Refine}\textsubscript{Contrast} over human references when translating into English. It demonstrates that our simple strategy enhances the naturalness of GPT outputs and that WMT human references could be less favourable than GPT outputs in some cases.

\subsection{Overall quality}
We also evaluate for general quality as a safeguard. In this setup, a source sentence and two translations are given to an evaluator who is fluent in both languages. They are asked to pick the translation with better quality or indicate a tie. We only evaluated two translation directions, English to and from Chinese, due to the limited availability of bilingual speakers. Similar to the previous evaluation, we compare \textit{Refine}\textsubscript{Contrast} against human references, as well as \textit{Refine}\textsubscript{Contrast} against \textit{Translate} separately. 

We report evaluator preferences in Figure~\ref{fig:human-evaluation} (right). It shows that GPT \textit{Refine} attains slightly better performance in zh$\rightarrow$en and similar performance in en$\rightarrow$zh when compared with human references. On the other hand, it is more favourable than GPT \textit{Translate} in terms of human judgements. Combining evaluation outcomes, we conclude that the refinement strategy could improve the target-side naturalness without undermining general quality.

\begin{table*}[th]
    \centering\small
    \input{tables/table_example.tex}
    \caption{German$\rightarrow$English and Chinese$\rightarrow$English examples showing rich lexical variations across translation strategies.}
    \label{tab:case_study}
\end{table*}

\section{Analysis and Discussions}
\subsection{Performance through iterations}

To investigate the behaviour of refinement strategies through different iterations, we plot BLEU, COMET\textsubscript{DA}, and COMET\textsubscript{QE} at different iterations in Figure~\ref{fig:trends} for four translation directions: en$\leftrightarrow$de and en$\leftrightarrow$zh. We find that \textit{Refine} and \textit{Refine}\textsubscript{Contrast} usually attain their best after undergoing more than one refinement iteration, showing superiority to one-off editing.\footnote{The first iteration is equivalent to a one-off translation editing using an LLM.} However, in almost all \textit{Paraphrase} experiments, scores decrease monotonically, indicating that semantics drift away as paraphrasing iterates. Moreover, \textit{Refine}\textsubscript{Random} results start low, gradually catch up, but never reach as high as \textit{Refine} or \textit{Refine}\textsubscript{Contrast}. This means that iterative refinement is indeed useful in fixing translations, but starting with a reasonable translation is also crucial for obtaining a strong result.

\subsection{Diverging automatic scores}
According to automatic string-based metrics, our queries deliver lower-quality translations through iterations, but COMET\textsubscript{DA} scores remain comparable and COMET\textsubscript{QE} scores mostly increase. We argue that the string-based metrics might not accurately indicate quality, but rather reflect text variations with respect to the reference. We further verified this via human evaluation that fluency and overall quality are not impacted.

In Table~\ref{tab:case_study} we show outputs from different strategies for a single source input, where a native speaker marked preference for \textit{Refine}\textsubscript{Contrast}. It illustrates that the word choice is diverse for both directions and specifically for Chinese$\rightarrow$English, there are substantial structural changes. The huge variety in expressions across translations can result in low BLEU with respect to human references, but without much change in meaning, for instance, as in Table~\ref{tab:automatic_results_gpt_main} where BLEU can decline up to one-third, but neural metric scores change little. In the field of MT, a leap in BLEU is usually associated with performance improvement; however, in our case, a drop cannot be simply interpreted as performance degradation. This can be attributed to the lexical and structural diversity in the refined translations.

\subsection{Human performance}
A human translator is deemed to be fluent in their native language, which intuitively is difficult for a model to compete with. In our human evaluation, GPT fluency can be as good or even better than reference translations---we offer two possible explanations. First, the WMT references might have been created by translators with varying expertise, which may not represent upper-bound human performance, especially when compared with advanced LLMs. More importantly, translations can exhibit awkwardness in word and syntax choices, potentially due to source language interference or ``shining through'' \cite{gellerstam,teich2003}.

\subsection{Relation to translationese}

Both human and machine translations might be more explicit, language-normalized, and simpler \cite{baker-corpus,koppel-ordan-2011-translationese}. On a broader scope, translationese is regarded as the distinct features in translations to include influences from both the source and target sides. Although MT normally learns from human translation data, researchers found that human and machine translation patterns do not fully overlap \cite{bizzoni-etal-2020-human}. While translationese occurs in translations inevitably, consumers could prefer translations that are more natural in their native language, provided that the semantics and utility are preserved.

From a narrow aspect, our method relates to machine translationese mitigation in terms of reducing unnaturalness and literalness, instead of focusing on state-of-the-art metric scores. It may be viable to create diverse translations through iterations, as we observe huge changes in BLEU scores. Measuring these using automatic metrics at the moment is challenging, especially given that most translation metrics are reference-based, where the reference can be translationese-prone in the first place. COMET\textsubscript{QE} might be more robust to this end.

\section{Related Work}\label{sec:appendix-related-work}
\subsection{Translation post-editing}
Closely related to our refinement prompting is automatic post-editing (APE), which trains a neural network to fix translation errors by learning from human correction data, that can be traced back to as early as \cite{knight-ape}. While it has shown advancements in statistical machine translation, it has been suspected to be less effective in the deep learning era due to original translations being high-quality and lack of post-editing data \cite{junczys-dowmunt-grundkiewicz-2018-ms,chatterjee-etal-2018-findings}. Whilst one way to facilitate this is more data provision \cite{chollampatt-etal-2020-automatic,ive-etal-2020-post}, our workaround utilizes a large language model, which possesses the post-editing capability without the need for specific training or fine-tuning. Furthermore, post-editing models might have limited power to alleviate awkwardness, because human editing data is collected from annotators who are usually instructed to not make style improvements \cite{ive-etal-2020-post}. Compared to APE, our method allows LLMs to re-generate an entirely different translation, which could escape the ``post-editese'' phenomenon, where Toral~\shortcite{toral-2019-post} demonstrated that human-edited machine translations still exhibit translationese features.

Some post-editing models do not rely on the source translation or human editing data \cite{simard-etal-2007-statistical}. For instance, Freitag~et~al.~\shortcite{freitag-etal-2019-ape} trained a post-editor solely on monolingual data by reconstructing the original text given its round-trip translation. In our work, we incorporate stronger natural language modelling into post-editing by employing LLMs. Other translation refinement research includes combining statistical and neural systems \cite{novak2016iterative,niehues-etal-2016-pre}, merging APE into the NMT framework \cite{pal-etal-2020-transference,chen-etal-2022-synchronous}, and debiasing translationese in the latent embedding space \cite{dutta-chowdhury-etal-2022-towards}. The iterative editing mechanism mostly lies in non-autoregressive translation, where each output token is independent of other target positions and iterative decoding enhances output quality \cite{lee-etal-2018-deterministic,gu-etal-2018-levenshtein,xu-carpuat-2021-editor}.

\subsection{Translation prompting with large language models}
Large language models have recently become highly effective tools for various NLP tasks  \cite{Radford2019Language,brown2020language,Chowdhery2022PaLM,ouyang2022training}. Nowadays, optimising LLMs directly for specific tasks becomes less important since they generalize to downstream tasks even without explicit supervision. With more parameters and training data, LLMs may offer stronger performance than dedicated translation or post-editing models. The method we use to elicit a response from GPT is zero-shot prompting \cite{brown2020language}, which means affixing a description to the original task input to form a query to the model. Researchers have benchmarked LLMs' capability to translate \cite{vilar-etal-2023-prompting,zhang2023prompting,jiao2023chatgpt,hendy2023good}, and to interpret translation quality \cite{kocmi2023large,lu2023error,xu2023instructscore}.

Among the recent papers on LLM translation prompting, we identify the following to be most relevant to us. Previous findings show that GPT produces less literal translations, especially for out-of-English translations \cite{raunak-etal-2023-gpts}, which to some extent stands in contrast with our later human evaluation results on naturalness and fluency. Raunak~et~al.~\shortcite{raunak2023leveraging} formalized post-editing as a chain-of-thought process \cite{wei2022chain} with GPT-4 and achieved promising results. Different from their focus, our work features the iterative refinement process as a means to enhance naturalness and fluency. Our work reveals that iterated refinement is better than one-off editing. The observed improvement, especially for into-English, may be attributed to the abundant English pre-training data available for LLMs. To the best of our knowledge, although the concept of iterative refinement is not new, ours is the pioneering paper in applying such strategies to LLMs for translation.

\section{Conclusion and Future Work}
We presented a simple way to leverage an LLM for translation refinement, which greatly helps fluency and naturalness. It is shown that our method maintains translation quality and introduces lexical and structural changes, especially for high-resource into-English translation. We have also discussed the potential of using our work to obtain diverse, fluent translations that are less translationese, as well as the limitation in automatic metrics to measure this.

On a broader note, this work connects to the concept of using LLMs to imitate collaborative translation refinement. Yet, it is important to acknowledge the high cost of running a multi-round LLM refinement. Future work can explore sentence-level refinement decisions to reduce cost.

\section*{Acknowledgement}
We express our gratitude to the reviewers of this paper for their detailed and invaluable feedback and suggestions. The work also benefited from discussions with Nikolay Bogoychev and Biao Zhang. We are grateful to Laurie Burchell, Ziqin Fang, Matthias Lindemann, and Jonas Waldendorf for their participation in the human evaluation.

This work is funded by UK Research and Innovation (UKRI) under the UK government’s Horizon Europe funding guarantee [grant number 10052546].

\bibliography{custom}
\bibliographystyle{eamt24}
\end{document}

%% file: tables/table_prompt_text.tex
\begin{tabular}{ll}
\toprule
\multicolumn{1}{c}{\textbf{Mode}} & \multicolumn{1}{c}{\textbf{Prompt}} \\
\midrule
\textit{Translate} & \texttt{Source: \source{\$\{source\}}} \\
& \texttt{Please give me a translation in \lang{\$\{lang\}} without any explanation.} \\
\cdashlinelr{1-2}
\textit{Refine}\textsubscript{} & \texttt{Source: \source{\$\{source\}}} \\
& \texttt{Translation: \translation{\$\{prev\_translation\}}} \\
& \texttt{Please give me a better \lang{\$\{lang\}} translation without any explanation.} \\
\cdashlinelr{1-2}
\textit{Refine}\textsubscript{Contrast} & \texttt{Source: \source{\$\{source\}}} \\
& \texttt{\textbf{Bad} translation: \translation{\$\{prev\_translation\}}} \\
& \texttt{Please give me a better \lang{\$\{lang\}} translation without any explanation.} \\
\cdashlinelr{1-2}
\textit{Refine}\textsubscript{Random} & \texttt{Source: \source{\$\{source\}}} \\
& \texttt{\textbf{Bad} translation: \randomtranslation{\$\{random\_target\}}} \textit{if first-round, else} \texttt{\translation{\$\{prev\_translation\}}} \\
& \texttt{Please give me a better \lang{\$\{lang\}} translation without any explanation.} \\
\cdashlinelr{1-2}
\textit{Paraphrase} & \texttt{Sentence: \translation{\$\{prev\_translation\}}} \\
& \texttt{Please give me a paraphrase in \lang{\$\{lang\}} without any explanation.} \\
\bottomrule
\end{tabular}

%% file: tables/table_automatic_results_gpt_previous.tex
\begin{tabular}{clc@{\hskip 3ex}c@{\hskip 1ex}c@{\hskip 0.25ex}c}
\toprule
         & & {BLEU} & {chrF++} & {COMET\textsubscript{DA}} & {COMET\textsubscript{QE}} \\
\midrule
    \multirow{6}{*}{\makecell{de\\$\downarrow$\\en}} & \textit{Reference\textsubscript{A}} & - & - & - & \textit{.0919} \\
\cdashlinelr{2-6}
    & Translate & \textbf{30.90} & \textbf{57.55} & \textbf{.8606} & .1128 \\
    & Refine & 23.14 & 51.91 & .8525 & .1116 \\
    & Refine\textsubscript{Contrast} & 22.88 & 52.47 & .8452 & \textbf{.1162} \\
    & Refine\textsubscript{Random} & 18.83 & 51.79 & .7777 & .0770 \\
    & Paraphrase & 11.01 & 40.05 & .8044 & .0919 \\
\midrule
    \multirow{6}{*}{\makecell{en\\$\downarrow$\\de}} & \textit{Reference\textsubscript{A}} & - & - & - & \textit{.1127} \\
\cdashlinelr{2-6}
    & Translate & \textbf{25.39} & \textbf{53.54} & .8427 & .1083 \\
    & Refine & 22.35 & 50.57 & \textbf{.8478} & \textbf{.1153} \\
    & Refine\textsubscript{Contrast} & 22.54 & 51.21 & .8211 & .0929 \\
    & Refine\textsubscript{Random} & 19.36 & 46.56 & .7906 & .0832 \\
    & Paraphrase & 13.60 & 43.54 & .8197 & .1006 \\
\midrule
    \multirow{6}{*}{\makecell{zh\\$\downarrow$\\en}} & \textit{Reference\textsubscript{A}} & - & - & - & \textit{.0708} \\
\cdashlinelr{2-6}
    & Translate & \textbf{25.64} & \textbf{53.74} & .8199 & .0867 \\
    & Refine & 20.26 & 49.06 & .8156 & .0921 \\
    & Refine\textsubscript{Contrast} & 24.81 & 51.77 & \textbf{.8538} & \textbf{.1132} \\
    & Refine\textsubscript{Random} & 24.24 & 47.11 & .8323 & .1022 \\
    & Paraphrase & 12.76 & 40.92 & .7931 & .0885 \\
\midrule
    \multirow{6}{*}{\makecell{en\\$\downarrow$\\zh}} & \textit{Reference\textsubscript{A}} & - & - & - & \textit{.0956} \\
\cdashlinelr{2-6}
    & Translate & \textbf{29.28} & \textbf{20.61} & .8300 & .0761 \\
    & Refine & 28.26 & 19.28 & \textbf{.8417} & .0870 \\
    & Refine\textsubscript{Contrast} & 29.28 & 19.69 & .8395 & \textbf{.0881} \\
    & Refine\textsubscript{Random} & 25.71 & 17.49 & .8126 & .0763 \\
    & Paraphrase & 21.95 & 17.14 & .8144 & .0716 \\
\bottomrule
\end{tabular}

%% file: tables/table_automatic_results_non_en_previous.tex
\begin{tabular}{clc@{\hskip 3ex}c@{\hskip 1ex}c@{\hskip 0.25ex}c}
\toprule
         & & {BLEU} & {chrF++} & {COMET\textsubscript{DA}} & {COMET\textsubscript{QE}} \\
\midrule
    \multirow{5}{*}{\makecell{de\\$\downarrow$\\fr}} & \textit{Reference} & - & - & - & \textit{.0772} \\
\cdashlinelr{2-6}
    & Translate & \textbf{36.25} & \textbf{59.50} & \textbf{.8395} & .0807 \\
    & Refine & 32.47 & 55.83 & .8353 & \textbf{.0851} \\
    & Refine\textsubscript{Contrast} & 33.12 & 56.37 & .8308 & .0805 \\
    & Paraphrase & 16.06 & 44.28 & .7937 & .0682 \\
\midrule
    \multirow{5}{*}{\makecell{en\\$\downarrow$\\ja}} & \textit{Reference} & - & - & - & \textit{.1345} \\
\cdashlinelr{2-6}
    & Translate & \textbf{23.00} & 25.89 & .8863 & .1255 \\
    & Refine & 22.63 & \textbf{27.30} & \textbf{.8941} & \textbf{.1305} \\
    & Refine\textsubscript{Contrast} & 22.82 & 26.71 & .8928 & .1282 \\
    & Paraphrase & 17.69 & 23.18 & .8592 & .1086 \\
\midrule
    \multirow{5}{*}{\makecell{uk\\$\downarrow$\\cs}} & \textit{Reference} & - & - & - & \textit{.1273} \\
\cdashlinelr{2-6}
    & Translate & \textbf{29.91} & \textbf{54.64} & \textbf{.9074} & .1173 \\
    & Refine & 28.60 & 53.06 & .9040 & \textbf{.1183} \\
    & Refine\textsubscript{Contrast} & 28.90 & 54.29 & .9036 & .1151 \\
    & Paraphrase & 13.59 & 40.04 & .8625 & .0969 \\
\bottomrule
\end{tabular}

%% file: tables/table_automatic_results_wmt_previous.tex
\begin{tabular}{clc@{\hskip 3ex}c@{\hskip 1ex}c@{\hskip 0.25ex}c}
\toprule
        & & {BLEU} & {chrF++} & {COMET\textsubscript{DA}} & {COMET\textsubscript{QE}} \\
\midrule
& \textit{Reference\textsubscript{A}} & - & - & - & \textit{.0919} \\

\cdashlinelr{1-6}
\multirow{4}{*}{\rotatebox[origin=c]{90}{Reference\textsubscript{B}}} & Submission & \textbf{30.05} & \textbf{56.00} & .8497 & .1050 \\
& Refine & 23.39 & 51.80 & .8527 & \textbf{.1123} \\
& Refine\textsubscript{Contrast} & 25.10 & 53.82 & \textbf{.8566} & .1116 \\
& Paraphrase & 12.52 & 41.03 & .8031 & .0894 \\

\cdashlinelr{1-6}
\multirow{4}{*}{\rotatebox[origin=c]{90}{Online\textsubscript{A}}} & Submission & \textbf{34.45} & \textbf{60.78} & \textbf{.8582} & .1061 \\
& Refine & 23.37 & 51.67 & .8494 & .1098 \\
& Refine\textsubscript{Contrast} & 25.14 & 52.84 & .8534 & \textbf{.1137} \\
& Paraphrase & 12.22 & 41.34 & .8097 & .0942 \\

\cdashlinelr{1-6}
\multirow{4}{*}{\rotatebox[origin=c]{90}{Online\textsubscript{Y}}} & Submission & \textbf{32.70} & \textbf{59.32} & .8500 & .0981 \\
& Refine & 22.92 & 50.85 & \textbf{.8522} & .1080 \\
& Refine\textsubscript{Contrast} & 24.40 & 53.32 & .8517 & \textbf{.1134} \\
& Paraphrase & 11.97 & 40.29 & .8054 & .0892\\

\cdashlinelr{1-6}
\multirow{4}{*}{\rotatebox[origin=c]{90}{Tencent}} & Submission & \textbf{35.35} & \textbf{61.28} & \textbf{.8584} & .1055 \\
& Refine & 23.75 & 52.16 & .8488 & .1095 \\
& Refine\textsubscript{Contrast} & 26.89 & 54.75 & .8553 & \textbf{.1116} \\
& Paraphrase & 12.43 & 41.35 & .8116 & .0947 \\

\cdashlinelr{1-6}
\multirow{4}{*}{\rotatebox[origin=c]{90}{Facebook}} & Submission & \textbf{34.67 }& \textbf{60.78} & \textbf{.8677} & \textbf{.1146} \\
& Refine & 22.97 & 51.05 & .8505 & .1113 \\
& Refine\textsubscript{Contrast} & 25.74 & 53.88 & .8548 & .1130 \\
& Paraphrase & 11.80 & 40.99 & .8099 & .0922 \\

\cdashlinelr{1-6}
\multirow{4}{*}{\rotatebox[origin=c]{90}{Edinburgh}} & Submission & \textbf{34.20} & \textbf{60.03} & \textbf{.8588} & .1087 \\
& Refine & 22.04 & 50.29 & .8496 & .1097 \\
& Refine\textsubscript{Contrast} & 25.24 & 52.87 & .8546 & \textbf{.1147} \\
& Paraphrase & 12.79 & 40.18 & .8067 & .0921 \\

\cdashlinelr{1-6}
\multirow{4}{*}{\rotatebox[origin=c]{90}{Huawei}} & Submission & \textbf{35.13} & \textbf{61.17} & \textbf{.8643} & \textbf{.1126} \\
& Refine & 22.24 & 50.82 & .8519 & .1097 \\
& Refine\textsubscript{Contrast} & 24.95 & 52.47 & .8560 & \textbf{.1124} \\
& Paraphrase & 12.20 & 40.74 & .8078 & .0909 \\
\bottomrule
\end{tabular}

%% file: plots/plot_source_free_refine_vs_translate_new.tex
\begin{tikzpicture}
\begin{axis}[
    xlabel={Fluency \& naturalness:\\Refine vs. Translate},
    xlabel style={align=center, text width=1\linewidth},
    width=5.75cm, height=4cm,
    ybar,
    ymax=50,ymin=0,
    bar width=0.85ex,
    enlarge y limits={0.0},
    enlarge x limits=0.12,
    symbolic x coords={
                    de$\rightarrow$en,
                    en$\rightarrow$de,
                    zh$\rightarrow$en,
                    en$\rightarrow$zh},
    tickwidth=0ex,
    xtick=data,
    xtick align=inside,
    xticklabel style={rotate=0, anchor=north},
    nodes near coords,
    nodes near coords style={font=\scriptsize, rotate=90, anchor=west},
    xticklabel style={font=\small},
    yticklabel style={font=\scriptsize},
    ]
\addplot[
        draw=black!75,
        fill=purple!30,
        text=black
] plot coordinates {
        (de$\rightarrow$en,3)
        (en$\rightarrow$de,23)
        (zh$\rightarrow$en,5)
        (en$\rightarrow$zh,17)
    };
\addplot[
        draw=black!75,
        fill=orange!30,
        text=black
] plot coordinates {
        (de$\rightarrow$en,28)
        (en$\rightarrow$de,15)
        (zh$\rightarrow$en,29)
        (en$\rightarrow$zh,23)
    };
\addplot[
        draw=black!75,
        fill=teal!30,
        text=black
] plot coordinates {
        (de$\rightarrow$en,19)
        (en$\rightarrow$de,12)
        (zh$\rightarrow$en,16)
        (en$\rightarrow$zh,10)
    };
\end{axis}
\end{tikzpicture}

%% file: plots/plot_source_free_refine_vs_ref_new.tex
\begin{tikzpicture}
\begin{axis}[
    xlabel={Fluency \& naturalness:\\Refine vs. Reference},
    xlabel style={align=center, text width=1\linewidth},
    width=5.75cm, height=4cm,
    ybar,
    ymax=50,ymin=0,
    bar width=0.85ex,
    enlarge y limits={0.0},
    enlarge x limits=0.12,
    symbolic x coords={
                    de$\rightarrow$en,
                    en$\rightarrow$de,
                    zh$\rightarrow$en,
                    en$\rightarrow$zh},
    tickwidth=0ex,
    xtick=data,
    xtick align=inside,
    xticklabel style={rotate=0, anchor=north},
    nodes near coords,
    nodes near coords style={font=\scriptsize, rotate=90, anchor=west},
    xticklabel style={font=\small},
    yticklabel style={font=\scriptsize},
    ]
\addplot[
        draw=black!75,
        fill=purple!30,
        text=black
] plot coordinates {
        (de$\rightarrow$en,1)
        (en$\rightarrow$de,15) 
        (zh$\rightarrow$en,4)
        (en$\rightarrow$zh,17)
    };
\addplot[
        draw=black!75,
        fill=orange!30,
        text=black
] plot coordinates {
        (de$\rightarrow$en,40)
        (en$\rightarrow$de,16) 
        (zh$\rightarrow$en,35)
        (en$\rightarrow$zh,8)
    };
\addplot[
        draw=black!75,
        fill=blue!30,
        text=black
] plot coordinates {
        (de$\rightarrow$en,9)
        (en$\rightarrow$de,19) 
        (zh$\rightarrow$en,11)
        (en$\rightarrow$zh,25)
    };
\end{axis}
\end{tikzpicture}

%% file: plots/plot_source_based_refine_vs_translate_new.tex
\begin{tikzpicture}
\begin{axis}[
    xlabel={Quality:\\Refine vs. Translate},
    xlabel style={align=center, text width=1\linewidth},
    width=3.75cm, height=4cm,
    ybar,
    ymax=50,ymin=0,
    bar width=0.85ex,
    enlarge y limits={0.0},
    enlarge x limits=0.45,
    symbolic x coords={
                    zh$\rightarrow$en,
                    en$\rightarrow$zh},
    tickwidth=0ex,
    xtick=data,
    xtick align=inside,
    xticklabel style={rotate=0, anchor=north},
    nodes near coords,
    nodes near coords style={font=\scriptsize, rotate=90, anchor=west},
    xticklabel style={font=\small},
    yticklabel style={font=\scriptsize},
    ]
\addplot[
        draw=black!75,
        fill=purple!30,
        text=black
] plot coordinates {
        (zh$\rightarrow$en,31)
        (en$\rightarrow$zh,25)
    };
\addplot[
        draw=black!75,
        fill=orange!30,
        text=black
] plot coordinates {
        (zh$\rightarrow$en,12)
        (en$\rightarrow$zh,14)
    };
\addplot[
        draw=black!75,
        fill=teal!30,
        text=black
] plot coordinates {
        (zh$\rightarrow$en,7)
        (en$\rightarrow$zh,11)
    };
\end{axis}
\end{tikzpicture}

%% file: plots/plot_source_based_refine_vs_ref_new.tex
\begin{tikzpicture}
\begin{axis}[
    xlabel={Quality:\\Refine vs. Reference},
    xlabel style={align=center, text width=1\linewidth},
    width=3.75cm, height=4cm,
    ybar,
    ymax=50,ymin=0,
    bar width=0.85ex,
    enlarge y limits={0.0},
    enlarge x limits=0.45,
    symbolic x coords={
                    zh$\rightarrow$en,
                    en$\rightarrow$zh},
    tickwidth=0ex,
    xtick=data,
    xtick align=inside,
    xticklabel style={rotate=0, anchor=north},
    nodes near coords,
    nodes near coords style={font=\scriptsize, rotate=90, anchor=west},
    xticklabel style={font=\small},
    yticklabel style={font=\scriptsize},
    ]
\addplot[
        draw=black!75,
        fill=purple!30,
        text=black
] plot coordinates {
        (zh$\rightarrow$en,17)
        (en$\rightarrow$zh,11)
    };
\addplot[
        draw=black!75,
        fill=orange!30,
        text=black
] plot coordinates {
        (zh$\rightarrow$en,20)
        (en$\rightarrow$zh,20)
    };
\addplot[
        draw=black!75,
        fill=blue!30,
        text=black
] plot coordinates {
        (zh$\rightarrow$en,13)
        (en$\rightarrow$zh,19)
    };
\end{axis}
\end{tikzpicture}

%% file: plots/plot_source_based_free_legend.tex
\begin{tikzpicture}
\begin{customlegend}[
    legend columns=-1,
    legend style={
            column sep=4ex,
            draw=none,
    },
    legend entries={
            \hspace{-3.5ex}Tie,
            \hspace{-3.5ex}Refine\textsubscript{Contrast},
            \hspace{-3.5ex}Translate,
            \hspace{-3.5ex}Reference,
    }
]
\addlegendimage{
        area legend,
        draw=black!75,
        fill=purple!30,
        text=black,
}
\addlegendimage{
        area legend,
        draw=black!75,
        fill=orange!30,
        text=black,
}
\addlegendimage{
        area legend,
        draw=black!75,
        fill=teal!30,
        text=black,
}
\addlegendimage{
        area legend,
        draw=black!75,
        fill=blue!30,
        text=black,
}
\end{customlegend}
\end{tikzpicture}

%% file: plots/plot_de_en_bleu.tex
\begin{tikzpicture}
\begin{axis}[
 xlabel=de$\rightarrow$en BLEU,
 width=0.33\textwidth,
 height=0.25\textwidth,
 xmin=1,
 xmax=4,
 xtick={1,2,3,4},
 legend cell align={left},
 yticklabel style={
        /pgf/number format/fixed,
        /pgf/number format/fixed zerofill,
        /pgf/number format/precision=0
 },
 every node near coord/.append style = {
        skip 0.
 },
 scaled y ticks=false,
 ]
  \addplot[blue,
           semithick,
           dashed,
           samples=4
           ] {30.8984};
  \addplot[mark=square,
           semithick,
           purple,
           samples=4
           ] coordinates {
            (1,22.5047)
            (2,23.1411)
            (3,22.0778)
            (4,21.9997) };
    \addplot[mark=asterisk, mark size=0.8ex,
           semithick,
           teal,
           samples=4
           ] coordinates {
            (1,23.3004)
            (2,22.9420)
            (3,22.5574)
            (4,22.8791) };
    \addplot[mark=triangle, mark size=0.7ex,
           semithick,
           violet,
           samples=4
           ] coordinates {
            (1,20.8662)
            (2,18.7156)
            (3,18.9195)
            (4,18.8281) };
  \addplot[mark=o,
           semithick,
           brown,
           samples=4
           ] coordinates {
            (1,11.0124)
            (2,11.1666)
            (3,7.9556)
            (4,8.9447) };
\end{axis}
\end{tikzpicture}

%% file: plots/plot_de_en_comet_da.tex
\begin{tikzpicture}
\begin{axis}[
 xlabel=de$\rightarrow$en COMET\textsubscript{DA},
 width=0.33\textwidth,height=0.25\textwidth,
 xmin=1,
 xmax=4,
 xtick={1,2,3,4},
 legend cell align={left},
 yticklabel style={
        /pgf/number format/fixed,
        /pgf/number format/fixed zerofill,
        /pgf/number format/precision=2
 },
 every node near coord/.append style = {
        skip 0.
 },
 scaled y ticks=false,
 ]
  \addplot[blue,
           semithick,
           dashed,
           samples=4
           ] {0.8606};
  \addplot[mark=square,
           semithick,
           purple,
           samples=4
           ] coordinates {
            (1,0.8446)
            (2,0.8525)
            (3,0.8470)
            (4,0.8486) };
    \addplot[mark=asterisk, mark size=0.8ex,
           semithick,
           teal,
           samples=4
           ] coordinates {
            (1,0.8454)
            (2,0.8476)
            (3,0.8457)
            (4,0.8452) };
    \addplot[mark=triangle, mark size=0.7ex,
           semithick,
           violet,
           samples=4
           ] coordinates {
            (1,0.7567)
            (2,0.7686)
            (3,0.7779)
            (4,0.7777) };
  \addplot[mark=o,
           semithick,
           brown,
           samples=4
           ] coordinates {
            (1,0.8044)
            (2,0.7922)
            (3,0.7801)
            (4,0.7738) };
\end{axis}
\end{tikzpicture}

%% file: plots/plot_de_en_comet_qe.tex
\begin{tikzpicture}
\begin{axis}[
 xlabel=de$\rightarrow$en COMET\textsubscript{QE},
 width=0.33\textwidth,height=0.25\textwidth,
 xmin=1,
 xmax=4,
 xtick={1,2,3,4},
 legend cell align={left},
 yticklabel style={
        /pgf/number format/fixed,
        /pgf/number format/fixed zerofill,
        /pgf/number format/precision=2
 },
 scaled y ticks=false,
 ]
  \addplot[red,
           semithick,
           densely dotted,
           samples=4
           ] {0.0919};
  \addplot[blue,
           semithick,
           dashed,
           samples=4
           ] {0.1128};
  \addplot[mark=square,
           semithick,
           purple,
           samples=4
           ] coordinates {
            (1,0.1061)
            (2,0.1116)
            (3,0.1087)
            (4,0.1085) };
  \addplot[mark=asterisk, mark size=0.8ex,
           semithick,
           teal,
           samples=4
           ] coordinates {
            (1,0.1116)
            (2,0.1133)
            (3,0.1146)
            (4,0.1162) };
    \addplot[mark=triangle, mark size=0.7ex,
           semithick,
           violet,
           samples=4
           ] coordinates {
            (1,0.0596)
            (2,0.0703)
            (3,0.0749)
            (4,0.0770) };
  \addplot[mark=o,
           semithick,
           brown,
           samples=4
           ] coordinates {
            (1,0.0885)
            (2,0.0835)
            (3,0.0824)
            (4,0.0786) };
\end{axis}
\end{tikzpicture}

%% file: plots/plot_en_de_bleu.tex
\begin{tikzpicture}
\begin{axis}[
 xlabel=en$\rightarrow$de BLEU,
 width=0.33\textwidth,height=0.25\textwidth,
 xmin=1,
 xmax=4,
 xtick={1,2,3,4},
 legend cell align={left},
 yticklabel style={
        /pgf/number format/fixed,
        /pgf/number format/fixed zerofill,
        /pgf/number format/precision=0
 },
 scaled y ticks=false,
 ]
  \addplot[blue,
           semithick,
           dashed,
           samples=4
           ] {25.3944};
  \addplot[mark=square,
           semithick,
           purple,
           samples=4
           ] coordinates {
            (1,22.5792)
            (2,22.3513)
            (3,22.8337)
            (4,22.4318) };
    \addplot[mark=asterisk, mark size=0.8ex,
           semithick,
           teal,
           samples=4
           ] coordinates {
            (1,22.7165)
            (2,22.6575)
            (3,22.5119)
            (4,22.5385) };
    \addplot[mark=triangle, mark size=0.7ex,
           semithick,
           violet,
           samples=4
           ] coordinates {
            (1,20.8662)
            (2,18.7156)
            (3,18.9195)
            (4,18.8281)};
  \addplot[mark=o,
           semithick,
           brown,
           samples=4
           ] coordinates {
            (1,13.6035)
            (2,11.2733)
            (3,9.3226)
            (4,8.4844) };
\end{axis}
\end{tikzpicture}

%% file: plots/plot_en_de_comet_da.tex
\begin{tikzpicture}
\begin{axis}[
 xlabel=en$\rightarrow$de COMET\textsubscript{DA},
 width=0.33\textwidth,height=0.25\textwidth,
 xmin=1,
 xmax=4,
 xtick={1,2,3,4},
 legend cell align={left},
 yticklabel style={
        /pgf/number format/fixed,
        /pgf/number format/fixed zerofill,
        /pgf/number format/precision=2
 },
 scaled y ticks=false,
 ]
  \addplot[blue,
           semithick,
           dashed,
           samples=4
           ] {0.8427};
  \addplot[mark=square,
           semithick,
           purple,
           samples=4
           ] coordinates {
            (1,0.8458)
            (2,0.8478)
            (3,0.8438)
            (4,0.8420) };
    \addplot[mark=asterisk, mark size=0.8ex,
           semithick,
           teal,
           samples=4
           ] coordinates {
            (1,0.8205)
            (2,0.8221)
            (3,0.8189)
            (4,0.8211) };
    \addplot[mark=triangle, mark size=0.7ex,
           semithick,
           violet,
           samples=4
           ] coordinates {
            (1,0.7567)
            (2,0.7686)
            (3,0.7779)
            (4,0.7777)};
  \addplot[mark=o,
           semithick,
           brown,
           samples=4
           ] coordinates {
            (1,0.8197)
            (2,0.7964)
            (3,0.8051)
            (4,0.7904) };
\end{axis}
\end{tikzpicture}

%% file: plots/plot_en_de_comet_qe.tex
\begin{tikzpicture}
\begin{axis}[
 xlabel=en$\rightarrow$de COMET\textsubscript{QE},
 width=0.33\textwidth,height=0.25\textwidth,
 xmin=1,
 xmax=4,
 xtick={1,2,3,4},
 legend cell align={left},
 yticklabel style={
        /pgf/number format/fixed,
        /pgf/number format/fixed zerofill,
        /pgf/number format/precision=2
 },
 scaled y ticks=false,
 ]
  \addplot[red,
           semithick,
           densely dotted,
           samples=4
           ] {0.1127};
  \addplot[blue,
           semithick,
           dashed,
           samples=4
           ] {0.1083};
  \addplot[mark=square,
           semithick,
           purple,
           samples=4
           ] coordinates {
            (1,0.1135)
            (2,0.1153)
            (3,0.1126)
            (4,0.1138) };
  \addplot[mark=asterisk, mark size=0.8ex,
           semithick,
           teal,
           samples=4
           ] coordinates {
            (1,0.0895)
            (2,0.0900)
            (3,0.0919)
            (4,0.0929) };
    \addplot[mark=triangle, mark size=0.7ex,
           semithick,
           violet,
           samples=4
           ] coordinates {
            (1,0.0596)
            (2,0.0703)
            (3,0.0749)
            (4,0.0770)};
  \addplot[mark=o,
           semithick,
           brown,
           samples=4
           ] coordinates {
            (1,0.1006)
            (2,0.0898)
            (3,0.0948)
            (4,0.0887) };
\end{axis}
\end{tikzpicture}

%% file: plots/plot_zh_en_bleu.tex
\begin{tikzpicture}
\begin{axis}[
 xlabel=zh$\rightarrow$en BLEU,
 width=0.33\textwidth,height=0.25\textwidth,
 xmin=1,
 xmax=4,
 xtick={1,2,3,4},
 legend cell align={left},
 yticklabel style={
        /pgf/number format/fixed,
        /pgf/number format/fixed zerofill,
        /pgf/number format/precision=0
 },
 scaled y ticks=false,
 ]
  \addplot[blue,
           semithick,
           dashed,
           samples=4
           ] {25.6413};
  \addplot[mark=square,
           semithick,
           purple,
           samples=4
           ] coordinates {
            (1,21.2118)
            (2,22.5516)
            (3,20.6860)
            (4,20.2551) };
    \addplot[mark=asterisk, mark size=0.8ex,
           semithick,
           teal,
           samples=4
           ] coordinates {
            (1,23.1012)
            (2,24.8076)
            (3,22.7389)
            (4,23.6289) };
    \addplot[mark=triangle, mark size=0.7ex,
           semithick,
           violet,
           samples=4
           ] coordinates {
            (1,25.4342)
            (2,22.1578)
            (3,22.3744)
            (4,24.2435)};
  \addplot[mark=o,
           semithick,
           brown,
           samples=4
           ] coordinates {
            (1,12.7614)
            (2,12.0358)
            (3,9.9643)
            (4,9.4380) };
\end{axis}
\end{tikzpicture}

%% file: plots/plot_zh_en_comet_da.tex
\begin{tikzpicture}
\begin{axis}[
 xlabel=zh$\rightarrow$en COMET\textsubscript{DA},
 width=0.33\textwidth,height=0.25\textwidth,
 ymin=0.76,
 ymax=0.86,
 ytick={0.77,0.80,0.83,0.86},
 xmin=1,
 xmax=4,
 xtick={1,2,3,4},
 legend cell align={left},
 yticklabel style={
        /pgf/number format/fixed,
        /pgf/number format/fixed zerofill,
        /pgf/number format/precision=2
 },
 scaled y ticks=false,
 ]
  \addplot[blue,
           semithick,
           dashed,
           samples=4
           ] {0.8199};
  \addplot[mark=square,
           semithick,
           purple,
           samples=4
           ] coordinates {
            (1,0.8145)
            (2,0.8167)
            (3,0.8157)
            (4,0.8156) };
    \addplot[mark=asterisk, mark size=0.8ex,
           semithick,
           teal,
           samples=4
           ] coordinates {
            (1,0.8477)
            (2,0.8538)
            (3,0.8470)
            (4,0.8498) };
    \addplot[mark=triangle, mark size=0.7ex,
           semithick,
           violet,
           samples=4
           ] coordinates {
            (1,0.7752)
            (2,0.8015)
            (3,0.8126)
            (4,0.8134)};
  \addplot[mark=o,
           semithick,
           brown,
           samples=4
           ] coordinates {
            (1,0.7931)
            (2,0.7835)
            (3,0.7759)
            (4,0.7685) };
\end{axis}
\end{tikzpicture}

%% file: plots/plot_zh_en_comet_qe.tex
\begin{tikzpicture}
\begin{axis}[
 xlabel=zh$\rightarrow$en COMET\textsubscript{QE},
 width=0.33\textwidth,height=0.25\textwidth,
 xmin=1,
 xmax=4,
 xtick={1,2,3,4},
 legend cell align={left},
 yticklabel style={
        /pgf/number format/fixed,
        /pgf/number format/fixed zerofill,
        /pgf/number format/precision=2
 },
 scaled y ticks=false,
 ]
  \addplot[red,
           semithick,
           densely dotted,
           samples=4
           ] {0.0708};
  \addplot[blue,
           semithick,
           dashed,
           samples=4
           ] {0.0867};
  \addplot[mark=square,
           semithick,
           purple,
           samples=4
           ] coordinates {
            (1,0.0891)
            (2,0.0906)
            (3,0.0879)
            (4,0.0921) };
  \addplot[mark=asterisk, mark size=0.8ex,
           semithick,
           teal,
           samples=4
           ] coordinates {
            (1,0.1086)
            (2,0.1132)
            (3,0.1080)
            (4,0.1102) };
    \addplot[mark=triangle, mark size=0.7ex,
           semithick,
           violet,
           samples=4
           ] coordinates {
            (1,0.0526)
            (2,0.0670)
            (3,0.0763)
            (4,0.0754)};
  \addplot[mark=o,
           semithick,
           brown,
           samples=4
           ] coordinates {
            (1,0.0885)
            (2,0.0835)
            (3,0.0824)
            (4,0.0786) };
\end{axis}
\end{tikzpicture}

%% file: plots/plot_en_zh_bleu.tex
\begin{tikzpicture}
\begin{axis}[
 xlabel=en$\rightarrow$zh BLEU,
 width=0.33\textwidth,height=0.25\textwidth,
 xmin=1,
 xmax=4,
 xtick={1,2,3,4},
 legend cell align={left},
 yticklabel style={
        /pgf/number format/fixed,
        /pgf/number format/fixed zerofill,
        /pgf/number format/precision=0
 },
 scaled y ticks=false,
 ]
  \addplot[blue,
           semithick,
           dashed,
           samples=4
           ] {29.2826};
  \addplot[mark=square,
           semithick,
           purple,
           samples=4
           ] coordinates {
            (1,28.8925)
            (2,27.3705)
            (3,28.2574)
            (4,27.8688) };
    \addplot[mark=asterisk, mark size=0.8ex,
           semithick,
           teal,
           samples=4
           ] coordinates {
            (1,29.9406)
            (2,29.2007)
            (3,28.4853)
            (4,29.2797) };
    \addplot[mark=triangle, mark size=0.7ex,
           semithick,
           violet,
           samples=4
           ] coordinates {
            (1,23.1202)
            (2,25.6792)
            (3,25.7082)
            (4,26.3583)};
  \addplot[mark=o,
           semithick,
           brown,
           samples=4
           ] coordinates {
            (1,23.4919)
            (2,21.9514)
            (3,19.6119)
            (4,17.8111) };
\end{axis}
\end{tikzpicture}

%% file: plots/plot_en_zh_comet_da.tex
\begin{tikzpicture}
\begin{axis}[
 xlabel=en$\rightarrow$zh COMET\textsubscript{DA},
 width=0.33\textwidth,height=0.25\textwidth,
 xmin=1,
 xmax=4,
 xtick={1,2,3,4},
 legend cell align={left},
 yticklabel style={
        /pgf/number format/fixed,
        /pgf/number format/fixed zerofill,
        /pgf/number format/precision=2
 },
 scaled y ticks=false,
 ]
  \addplot[blue,
           semithick,
           dashed,
           samples=4
           ] {0.8300};
  \addplot[mark=square,
           semithick,
           purple,
           samples=4
           ] coordinates {
            (1,0.8332)
            (2,0.8348)
            (3,0.8417)
            (4,0.8383) };
    \addplot[mark=asterisk, mark size=0.8ex,
           semithick,
           teal,
           samples=4
           ] coordinates {
            (1,0.8394)
            (2,0.8372)
            (3,0.8392)
            (4,0.8395) };
    \addplot[mark=triangle, mark size=0.7ex,
           semithick,
           violet,
           samples=4
           ] coordinates {
            (1,0.7752)
            (2,0.8015)
            (3,0.8126)
            (4,0.8134)};
  \addplot[mark=o,
           semithick,
           brown,
           samples=4
           ] coordinates {
            (1,0.8205)
            (2,0.8144)
            (3,0.8073)
            (4,0.8063) };
\end{axis}
\end{tikzpicture}

%% file: plots/plot_en_zh_comet_qe.tex
\begin{tikzpicture}
\begin{axis}[
 xlabel=en$\rightarrow$zh COMET\textsubscript{QE},
 width=0.33\textwidth,height=0.25\textwidth,
 ymax=0.10,
 xmin=1,
 xmax=4,
 xtick={1,2,3,4},
 legend cell align={left},
 yticklabel style={
        /pgf/number format/fixed,
        /pgf/number format/fixed zerofill,
        /pgf/number format/precision=2
 },
 scaled y ticks=false,
 ]
  \addplot[red,
           semithick,
           densely dotted,
           samples=4
           ] {0.0956};
  \addplot[blue,
           semithick,
           dashed,
           samples=4
           ] {0.0761};
  \addplot[mark=square,
           semithick,
           purple,
           samples=4
           ] coordinates {
            (1,0.0804)
            (2,0.0870)
            (3,0.0879)
            (4,0.0861) };
  \addplot[mark=asterisk, mark size=0.8ex,
           semithick,
           teal,
           samples=4
           ] coordinates {
            (1,0.0833)
            (2,0.0841)
            (3,0.0848)
            (4,0.0881) };
    \addplot[mark=triangle, mark size=0.7ex,
           semithick,
           violet,
           samples=4
           ] coordinates {
            (1,0.0526)
            (2,0.0670)
            (3,0.0763)
            (4,0.0754)};
  \addplot[mark=o,
           semithick,
           brown,
           samples=4
           ] coordinates {
            (1,0.0715)
            (2,0.0716)
            (3,0.0681)
            (4,0.0667) };
\end{axis}
\end{tikzpicture}

%% file: plots/plot_legend_scores.tex
\begin{tikzpicture}
\begin{customlegend}[
legend columns=-1,
legend style={
    draw=none,
    column sep=3ex},
    legend entries={
    \hspace{-3ex}Reference\textsubscript{A},
    \hspace{-3ex}Translate,
    \hspace{-3ex}Refine,
    \hspace{-3ex}Refine\textsubscript{Contrast},
    \hspace{-3ex}Refine\textsubscript{Random},
    \hspace{-3ex}Paraphrase
    }
]
\addlegendimage{densely dotted, thick, red}
\addlegendimage{dashed, thick, blue}
\addlegendimage{mark=square, thick, purple}
\addlegendimage{mark=asterisk, mark size=0.8ex, thick, teal}
\addlegendimage{mark=triangle, mark size=0.7ex, thick, violet}
\addlegendimage{mark=o, thick, brown}
\end{customlegend}
\end{tikzpicture}

%% file: tables/table_example.tex
\begin{tabularx}{\textwidth}{lX}
\toprule
   Source & Der 17-J\"{a}hrige \textcolor{orange}{floh} \textcolor{magenta}{zun\"{a}chst} vom \textcolor{teal}{Tatort}, seine Personalien konnten aber im Nachhinein \textcolor{cyan}{ermittelt werden}. \\
   Reference & The 17 year-old \textcolor{magenta}{proceeded} to \textcolor{orange}{flee} \textcolor{teal}{the crime scene}, however, his personal details could \textcolor{cyan}{be retrieved} later. \\
   Translate & The 17-year-old \textcolor{magenta}{initially} \textcolor{orange}{fled} from \textcolor{teal}{the crime scene}, but his personal information \textcolor{cyan}{was} later \textcolor{cyan}{determined}. \\
   Refine\textsubscript{Contrast} & The 17-year-old \textcolor{magenta}{initially} \textcolor{orange}{fled} from \textcolor{teal}{the scene of the crime}, but his personal details could later \textcolor{cyan}{be identified}. \\
   Paraphrase & \textcolor{magenta}{At first}, the 17-year-old \textcolor{orange}{ran away} from \textcolor{teal}{where the crime occurred}, but eventually, the authorities were able to \textcolor{cyan}{identify} him by his personal details. \\   
   \midrule
   Source & \begin{CJK*}{UTF8}{gbsn}新\textcolor{orange}{法令}\textcolor{teal}{规定}，坎帕尼亚大区自即日起室内公共场所必须戴口罩，\textcolor{magenta}{违者}最高可处以1000欧元罚金。\end{CJK*}\\
   Reference & \textcolor{teal}{According to} a new \textcolor{orange}{decree}, people must wear masks in indoor public places in Campania from now on, and \textcolor{magenta}{offenders} can be fined up to 1,000 euros. \\
   Translate & A new \textcolor{orange}{regulation} \textcolor{teal}{stipulates} that in Campania, indoor public places must wear masks. \textcolor{magenta}{Violators} can be fined up to 1000 euros. \\
   Refine\textsubscript{Contrast} & A new \textcolor{orange}{regulation} \textcolor{teal}{states} that in the Campania region, masks must be worn in indoor public places, with a maximum fine of 1000 euros for \textcolor{magenta}{those who violate the rule}. \\
   Paraphrase & A new \textcolor{orange}{rule} in Campania \textcolor{teal}{requires} people to wear masks in indoor public places, and \textcolor{magenta}{those who don't follow this rule} may be charged up to 1000 euros. \\
   \bottomrule
\end{tabularx}

%% file: eamt24.bbl
\begin{thebibliography}{}

\bibitem[\protect\citename{Agrawal \bgroup et al.\egroup }2023]{agrawal-etal-2023-context}
Agrawal, Sweta, Chunting Zhou, Mike Lewis, Luke Zettlemoyer, and Marjan Ghazvininejad.
\newblock 2023.
\newblock In-context examples selection for machine translation.
\newblock In {\em Findings of the Association for Computational Linguistics: ACL 2023}.

\bibitem[\protect\citename{Bahdanau \bgroup et al.\egroup }2015]{Bahdanau2015}
Bahdanau, Dzmitry, Kyunghyun Cho, and Yoshua Bengio.
\newblock 2015.
\newblock Neural machine translation by jointly learning to align and translate.
\newblock In {\em 3rd International Conference on Learning Representations}.

\bibitem[\protect\citename{Baker}1996]{baker-corpus}
Baker, Mona, 1996.
\newblock {\em Corpus-based Translation Studies: The Challenges that Lie Ahead}.
\newblock Benjamins Translation Library. John Benjamins Publishing Company.

\bibitem[\protect\citename{Bizzoni \bgroup et al.\egroup }2020]{bizzoni-etal-2020-human}
Bizzoni, Yuri, Tom~S Juzek, Cristina Espa{\~n}a-Bonet, Koel Dutta~Chowdhury, Josef van Genabith, and Elke Teich.
\newblock 2020.
\newblock How human is machine translationese? comparing human and machine translations of text and speech.
\newblock In {\em Proceedings of the 17th International Conference on Spoken Language Translation}.

\bibitem[\protect\citename{Brown \bgroup et al.\egroup }2020]{brown2020language}
Brown, Tom, Benjamin Mann, Nick Ryder, Melanie Subbiah, Jared~D Kaplan, Prafulla Dhariwal, Arvind Neelakantan, Pranav Shyam, Girish Sastry, Amanda Askell, et~al.
\newblock 2020.
\newblock Language models are few-shot learners.
\newblock In {\em Advances in Neural Information Processing Systems}.

\bibitem[\protect\citename{Chatterjee \bgroup et al.\egroup }2018]{chatterjee-etal-2018-findings}
Chatterjee, Rajen, Matteo Negri, Raphael Rubino, and Marco Turchi.
\newblock 2018.
\newblock Findings of the {WMT} 2018 shared task on automatic post-editing.
\newblock In {\em Proceedings of the Third Conference on Machine Translation}.

\bibitem[\protect\citename{Chen \bgroup et al.\egroup }2021]{chen-etal-2021-university}
Chen, Pinzhen, Jind{\v{r}}ich Helcl, Ulrich Germann, Laurie Burchell, Nikolay Bogoychev, Antonio~Valerio Miceli~Barone, Jonas Waldendorf, Alexandra Birch, and Kenneth Heafield.
\newblock 2021.
\newblock The {U}niversity of {E}dinburgh{'}s {E}nglish-{G}erman and {E}nglish-{H}ausa submissions to the {WMT}21 news translation task.
\newblock In {\em Proceedings of the Sixth Conference on Machine Translation}.

\bibitem[\protect\citename{Chen \bgroup et al.\egroup }2022]{chen-etal-2022-synchronous}
Chen, Kehai, Masao Utiyama, Eiichiro Sumita, Rui Wang, and Min Zhang.
\newblock 2022.
\newblock Synchronous refinement for neural machine translation.
\newblock In {\em Findings of the Association for Computational Linguistics: ACL 2022}.

\bibitem[\protect\citename{Chollampatt \bgroup et al.\egroup }2020]{chollampatt-etal-2020-automatic}
Chollampatt, Shamil, Raymond~Hendy Susanto, Liling Tan, and Ewa Szymanska.
\newblock 2020.
\newblock Can automatic post-editing improve {NMT}?
\newblock In {\em Proceedings of the 2020 Conference on Empirical Methods in Natural Language Processing}.

\bibitem[\protect\citename{Chowdhery \bgroup et al.\egroup }2022]{Chowdhery2022PaLM}
Chowdhery, Aakanksha, Sharan Narang, Jacob Devlin, Maarten Bosma, Gaurav Mishra, Adam Roberts, Paul Barham, Hyung~Won Chung, Charles Sutton, Sebastian Gehrmann, et~al.
\newblock 2022.
\newblock {PaLM}: Scaling language modeling with pathways.
\newblock {\em arXiv preprint}.

\bibitem[\protect\citename{Dutta~Chowdhury \bgroup et al.\egroup }2022]{dutta-chowdhury-etal-2022-towards}
Dutta~Chowdhury, Koel, Rricha Jalota, Cristina Espa{\~n}a-Bonet, and Josef Genabith.
\newblock 2022.
\newblock Towards debiasing translation artifacts.
\newblock In {\em Proceedings of the 2022 Conference of the North American Chapter of the Association for Computational Linguistics: Human Language Technologies}.

\bibitem[\protect\citename{Farhad \bgroup et al.\egroup }2021]{akhbardeh-etal-2021-findings}
Farhad, Akhbardeh, Arkhangorodsky Arkady, Biesialska Magdalena, Bojar Ond{\v{r}}ej, Chatterjee Rajen, Chaudhary Vishrav, Marta~R Costa-jussa, Espa{\~n}a-Bonet Cristina, Fan Angela, Federmann Christian, et~al.
\newblock 2021.
\newblock Findings of the 2021 conference on machine translation ({WMT}21).
\newblock In {\em Proceedings of the Sixth Conference on Machine Translation}.

\bibitem[\protect\citename{Freitag \bgroup et al.\egroup }2019]{freitag-etal-2019-ape}
Freitag, Markus, Isaac Caswell, and Scott Roy.
\newblock 2019.
\newblock {APE} at scale and its implications on {MT} evaluation biases.
\newblock In {\em Proceedings of the Fourth Conference on Machine Translation}.

\bibitem[\protect\citename{Freitag \bgroup et al.\egroup }2022]{freitag-etal-2022-results}
Freitag, Markus, Ricardo Rei, Nitika Mathur, Chi-kiu Lo, Craig Stewart, Eleftherios Avramidis, Tom Kocmi, George Foster, Alon Lavie, and Andr{\'e} F.~T. Martins.
\newblock 2022.
\newblock Results of {WMT}22 metrics shared task: Stop using {BLEU} {--} neural metrics are better and more robust.
\newblock In {\em Proceedings of the Seventh Conference on Machine Translation}.

\bibitem[\protect\citename{Gellerstam}1986]{gellerstam}
Gellerstam, Martin.
\newblock 1986.
\newblock Translationese in {Swedish} novels translated from {English}.
\newblock In {\em Translation studies in Scandinavia: Proceedings from the Scandinavian Symposium on Translation Theory II}. CWK Gleerup.

\bibitem[\protect\citename{Gu \bgroup et al.\egroup }2019]{gu-etal-2018-levenshtein}
Gu, Jiatao, Changhan Wang, and Junbo Zhao.
\newblock 2019.
\newblock Levenshtein transformer.
\newblock In {\em Advances in Neural Information Processing Systems}.

\bibitem[\protect\citename{Hendy \bgroup et al.\egroup }2023]{hendy2023good}
Hendy, Amr, Mohamed Abdelrehim, Amr Sharaf, Vikas Raunak, Mohamed Gabr, Hitokazu Matsushita, Young~Jin Kim, Mohamed Afify, and Hany~Hassan Awadalla.
\newblock 2023.
\newblock How good are {GPT} models at machine translation? a comprehensive evaluation.
\newblock {\em arXiv preprint}.

\bibitem[\protect\citename{Ive \bgroup et al.\egroup }2020]{ive-etal-2020-post}
Ive, Julia, Lucia Specia, Sara Szoc, Tom Vanallemeersch, Joachim Van~den Bogaert, Eduardo Farah, Christine Maroti, Artur Ventura, and Maxim Khalilov.
\newblock 2020.
\newblock A post-editing dataset in the legal domain: Do we underestimate neural machine translation quality?
\newblock In {\em Proceedings of the Twelfth Language Resources and Evaluation Conference}.

\bibitem[\protect\citename{Jiao \bgroup et al.\egroup }2023]{jiao2023chatgpt}
Jiao, Wenxiang, Wenxuan Wang, Jen tse Huang, Xing Wang, and Zhaopeng Tu.
\newblock 2023.
\newblock Is {ChatGPT} a good translator? {Yes} with {GPT-4} as the engine.
\newblock {\em arXiv preprint}.

\bibitem[\protect\citename{Junczys-Dowmunt and Grundkiewicz}2018]{junczys-dowmunt-grundkiewicz-2018-ms}
Junczys-Dowmunt, Marcin and Roman Grundkiewicz.
\newblock 2018.
\newblock {MS}-{UE}din submission to the {WMT}2018 {APE} shared task: Dual-source transformer for automatic post-editing.
\newblock In {\em Proceedings of the Third Conference on Machine Translation}.

\bibitem[\protect\citename{Kaplan \bgroup et al.\egroup }2020]{kaplan2020scaling}
Kaplan, Jared, Sam McCandlish, Tom Henighan, Tom~B. Brown, Benjamin Chess, Rewon Child, Scott Gray, Alec Radford, Jeffrey Wu, and Dario Amodei.
\newblock 2020.
\newblock Scaling laws for neural language models.
\newblock {\em arXiv preprint}.

\bibitem[\protect\citename{Knight and Chander}1994]{knight-ape}
Knight, Kevin and Ishwar Chander.
\newblock 1994.
\newblock Automated postediting of documents.
\newblock In {\em Proceedings of the Twelfth AAAI National Conference on Artificial Intelligence}.

\bibitem[\protect\citename{Kocmi and Federmann}2023]{kocmi2023large}
Kocmi, Tom and Christian Federmann.
\newblock 2023.
\newblock Large language models are state-of-the-art evaluators of translation quality.
\newblock {\em arXiv preprint}.

\bibitem[\protect\citename{Kocmi \bgroup et al.\egroup }2022]{kocmi-etal-2022-findings}
Kocmi, Tom, Rachel Bawden, Ond{\v{r}}ej Bojar, Anton Dvorkovich, Christian Federmann, Mark Fishel, Thamme Gowda, Yvette Graham, Roman Grundkiewicz, Barry Haddow, et~al.
\newblock 2022.
\newblock Findings of the 2022 conference on machine translation ({WMT}22).
\newblock In {\em Proceedings of the Seventh Conference on Machine Translation}.

\bibitem[\protect\citename{Koppel and Ordan}2011]{koppel-ordan-2011-translationese}
Koppel, Moshe and Noam Ordan.
\newblock 2011.
\newblock Translationese and its dialects.
\newblock In {\em Proceedings of the 49th Annual Meeting of the Association for Computational Linguistics: Human Language Technologies}.

\bibitem[\protect\citename{Lee \bgroup et al.\egroup }2018]{lee-etal-2018-deterministic}
Lee, Jason, Elman Mansimov, and Kyunghyun Cho.
\newblock 2018.
\newblock Deterministic non-autoregressive neural sequence modeling by iterative refinement.
\newblock In {\em Proceedings of the 2018 Conference on Empirical Methods in Natural Language Processing}.

\bibitem[\protect\citename{Lembersky \bgroup et al.\egroup }2012]{Lembersky-lm}
Lembersky, Gennadi, Noam Ordan, and Shuly Wintner.
\newblock 2012.
\newblock {Language Models for Machine Translation: Original vs. Translated Texts }.
\newblock {\em Computational Linguistics}.

\bibitem[\protect\citename{Lu \bgroup et al.\egroup }2023]{lu2023error}
Lu, Qingyu, Baopu Qiu, Liang Ding, Liping Xie, and Dacheng Tao.
\newblock 2023.
\newblock Error analysis prompting enables human-like translation evaluation in large language models: A case study on {ChatGPT}.
\newblock {\em arXiv preprint}.

\bibitem[\protect\citename{Niehues \bgroup et al.\egroup }2016]{niehues-etal-2016-pre}
Niehues, Jan, Eunah Cho, Thanh-Le Ha, and Alex Waibel.
\newblock 2016.
\newblock Pre-translation for neural machine translation.
\newblock In {\em Proceedings of the 26th International Conference on Computational Linguistics}.

\bibitem[\protect\citename{Novak \bgroup et al.\egroup }2016]{novak2016iterative}
Novak, Roman, Michael Auli, and David Grangier.
\newblock 2016.
\newblock Iterative refinement for machine translation.
\newblock {\em arXiv preprint}.

\bibitem[\protect\citename{Ouyang \bgroup et al.\egroup }2022]{ouyang2022training}
Ouyang, Long, Jeffrey Wu, Xu~Jiang, Diogo Almeida, Carroll Wainwright, Pamela Mishkin, Chong Zhang, Sandhini Agarwal, Katarina Slama, Alex Ray, et~al.
\newblock 2022.
\newblock Training language models to follow instructions with human feedback.
\newblock In {\em Advances in Neural Information Processing Systems}.

\bibitem[\protect\citename{Pal \bgroup et al.\egroup }2020]{pal-etal-2020-transference}
Pal, Santanu, Hongfei Xu, Nico Herbig, Sudip~Kumar Naskar, Antonio Kr{\"u}ger, and Josef van Genabith.
\newblock 2020.
\newblock The transference architecture for automatic post-editing.
\newblock In {\em Proceedings of the 28th International Conference on Computational Linguistics}.

\bibitem[\protect\citename{Papineni \bgroup et al.\egroup }2002]{papineni-etal-2002-bleu}
Papineni, Kishore, Salim Roukos, Todd Ward, and Wei-Jing Zhu.
\newblock 2002.
\newblock {BLEU}: a method for automatic evaluation of machine translation.
\newblock In {\em Proceedings of the 40th Annual Meeting of the Association for Computational Linguistics}.

\bibitem[\protect\citename{Popovi{\'c}}2017]{popovic-2017-chrf}
Popovi{\'c}, Maja.
\newblock 2017.
\newblock chr{F}++: words helping character n-grams.
\newblock In {\em Proceedings of the Second Conference on Machine Translation}.

\bibitem[\protect\citename{Radford \bgroup et al.\egroup }2019]{Radford2019Language}
Radford, Alec, Jeff Wu, Rewon Child, David Luan, Dario Amodei, and Ilya Sutskever.
\newblock 2019.
\newblock Language models are unsupervised multitask learners.
\newblock openai.com.

\bibitem[\protect\citename{Raunak \bgroup et al.\egroup }2023a]{raunak-etal-2023-gpts}
Raunak, Vikas, Arul Menezes, Matt Post, and Hany Hassan.
\newblock 2023a.
\newblock Do {GPT}s produce less literal translations?
\newblock In {\em Proceedings of the 61st Annual Meeting of the Association for Computational Linguistics}.

\bibitem[\protect\citename{Raunak \bgroup et al.\egroup }2023b]{raunak2023leveraging}
Raunak, Vikas, Amr Sharaf, Hany~Hassan Awadallah, and Arul Menezes.
\newblock 2023b.
\newblock Leveraging {GPT-4} for automatic translation post-editing.
\newblock {\em arXiv preprint}.

\bibitem[\protect\citename{Rei \bgroup et al.\egroup }2020]{rei-etal-2020-comet}
Rei, Ricardo, Craig Stewart, Ana~C Farinha, and Alon Lavie.
\newblock 2020.
\newblock {COMET}: A neural framework for {MT} evaluation.
\newblock In {\em Proceedings of the 2020 Conference on Empirical Methods in Natural Language Processing}.

\bibitem[\protect\citename{Simard \bgroup et al.\egroup }2007]{simard-etal-2007-statistical}
Simard, Michel, Cyril Goutte, and Pierre Isabelle.
\newblock 2007.
\newblock Statistical phrase-based post-editing.
\newblock In {\em Human Language Technologies 2007: The Conference of the North {A}merican Chapter of the Association for Computational Linguistics}.

\bibitem[\protect\citename{Teich}2003]{teich2003}
Teich, Elke.
\newblock 2003.
\newblock {\em Cross-Linguistic Variation in System and Text: A Methodology for the Investigation of Translations and Comparable Texts}.
\newblock De Gruyter Mouton.

\bibitem[\protect\citename{Toral}2019]{toral-2019-post}
Toral, Antonio.
\newblock 2019.
\newblock Post-editese: An exacerbated translationese.
\newblock In {\em Proceedings of Machine Translation Summit XVII}.

\bibitem[\protect\citename{Tran \bgroup et al.\egroup }2021]{tran-etal-2021-facebook}
Tran, Chau, Shruti Bhosale, James Cross, Philipp Koehn, Sergey Edunov, and Angela Fan.
\newblock 2021.
\newblock {F}acebook {AI}{'}s {WMT}21 news translation task submission.
\newblock In {\em Proceedings of the Sixth Conference on Machine Translation}.

\bibitem[\protect\citename{Vaswani \bgroup et al.\egroup }2017]{vaswani2017}
Vaswani, Ashish, Noam Shazeer, Niki Parmar, Jakob Uszkoreit, Llion Jones, Aidan~N Gomez, \L~ukasz Kaiser, and Illia Polosukhin.
\newblock 2017.
\newblock Attention is all you need.
\newblock In {\em Advances in Neural Information Processing Systems}.

\bibitem[\protect\citename{Vilar \bgroup et al.\egroup }2023]{vilar-etal-2023-prompting}
Vilar, David, Markus Freitag, Colin Cherry, Jiaming Luo, Viresh Ratnakar, and George Foster.
\newblock 2023.
\newblock Prompting {P}a{LM} for translation: Assessing strategies and performance.
\newblock In {\em Proceedings of the 61st Annual Meeting of the Association for Computational Linguistics}.

\bibitem[\protect\citename{Wang \bgroup et al.\egroup }2021]{wang-etal-2021-tencent}
Wang, Longyue, Mu~Li, Fangxu Liu, Shuming Shi, Zhaopeng Tu, Xing Wang, Shuangzhi Wu, Jiali Zeng, and Wen Zhang.
\newblock 2021.
\newblock Tencent translation system for the {WMT}21 news translation task.
\newblock In {\em Proceedings of the Sixth Conference on Machine Translation}.

\bibitem[\protect\citename{Wei \bgroup et al.\egroup }2021]{wei-etal-2021-hw}
Wei, Daimeng, Zongyao Li, Zhanglin Wu, Zhengzhe Yu, Xiaoyu Chen, Hengchao Shang, Jiaxin Guo, Minghan Wang, Lizhi Lei, Min Zhang, et~al.
\newblock 2021.
\newblock {HW}-{TSC}{'}s participation in the {WMT} 2021 news translation shared task.
\newblock In {\em Proceedings of the Sixth Conference on Machine Translation}.

\bibitem[\protect\citename{Wei \bgroup et al.\egroup }2022]{wei2022chain}
Wei, Jason, Xuezhi Wang, Dale Schuurmans, Maarten Bosma, Brian Ichter, Fei Xia, Ed~H. Chi, Quoc~V Le, and Denny Zhou.
\newblock 2022.
\newblock Chain of thought prompting elicits reasoning in large language models.
\newblock In {\em Advances in Neural Information Processing Systems}.

\bibitem[\protect\citename{Xu and Carpuat}2021]{xu-carpuat-2021-editor}
Xu, Weijia and Marine Carpuat.
\newblock 2021.
\newblock {EDITOR}: An edit-based transformer with repositioning for neural machine translation with soft lexical constraints.
\newblock {\em Transactions of the Association for Computational Linguistics}.

\bibitem[\protect\citename{Xu \bgroup et al.\egroup }2023]{xu2023instructscore}
Xu, Wenda, Danqing Wang, Liangming Pan, Zhenqiao Song, Markus Freitag, William~Yang Wang, and Lei Li.
\newblock 2023.
\newblock {INSTRUCTSCORE}: Towards explainable text generation evaluation with automatic feedback.
\newblock {\em arXiv preprint}.

\bibitem[\protect\citename{Zhang \bgroup et al.\egroup }2023]{zhang2023prompting}
Zhang, Biao, Barry Haddow, and Alexandra Birch.
\newblock 2023.
\newblock Prompting large language model for machine translation: A case study.
\newblock In {\em Proceedings of the 40th International Conference on Machine Learning}.

\end{thebibliography}
